\documentclass[letterpaper, 10 pt, conference]{ieeeconf}  

\IEEEoverridecommandlockouts                              

\usepackage{hyperref}       
\usepackage{url}            
\usepackage{booktabs}       
\usepackage{amsfonts}       
\usepackage{nicefrac}       
\usepackage{microtype}      

\newcommand{\RNum}[1]{\uppercase\expandafter{\romannumeral #1\relax}}

\usepackage{graphicx}
\usepackage{subfigure}

\usepackage{amsmath}

\usepackage{float}

\title{\LARGE \bf
AU-Expression Knowledge Constrained Representation Learning \\ 
for Facial Expression Recognition
}

\author{Tao Pu$^{1}$, Tianshui Chen$^{2,*}$, Yuan Xie$^{2}$, Hefeng Wu$^{1}$, and Liang Lin$^{1,2}$
\thanks{*Tianshui Chen is the corresponding author}
\thanks{$^{1}$Tao Pu, Tianshui Chen, Hefeng Wu, and Liang Lin are with Human Cyber Physical Intelligence Integration Lab, School of Data and Computer Science, Sun Yat-sen University, Guangzhou, China
        {\tt\small \{putao537, tianshuichen, wuhefeng\}@gmail.com}}%
\thanks{$^{2}$Tianshui Chen, Yuan Xie, and Liang Lin are with DarkMatter AI Research, Guangzhou, China.
        {\tt\small \{tianshuichen, phoenixsysu\}@gmail.com}}
}

\begin{document}

\maketitle
\thispagestyle{empty}
\pagestyle{empty}
\begin{abstract}

Recognizing human emotion/expressions automatically is quite an expected ability for intelligent robotics, as it can promote better communication and cooperation with humans. Current deep-learning-based algorithms may achieve impressive performance in some lab-controlled environments, but they always fail to recognize the expressions accurately for the uncontrolled in-the-wild situation. Fortunately, facial action units (AU) describe subtle facial behaviors, and they can help distinguish uncertain and ambiguous expressions. In this work, we explore the correlations among the action units and facial expressions, and devise an AU-Expression Knowledge Constrained Representation Learning (AUE-CRL) framework to learn the AU representations without AU annotations and adaptively use representations to facilitate facial expression recognition. Specifically, it leverages AU-expression correlations to guide the learning of the AU classifiers, and thus it can obtain AU representations without incurring any AU annotations. Then, it introduces a knowledge-guided attention mechanism that mines useful AU representations under the constraint of AU-expression correlations. In this way, the framework can capture local discriminative and complementary features to enhance facial representation for facial expression recognition. We conduct experiments on the challenging uncontrolled datasets to demonstrate the superiority of the proposed framework over current state-of-the-art methods. Codes and trained models are available at https://github.com/HCPLab-SYSU/AUE-CRL.

\end{abstract}

\section{Introduction}

Facial expression recognition (FER) is essential for intelligent robotics because it can help the robotics to understand human emotions and behaviors. Basically, this task aims to classify basic (e.g., happy, angry) or compound (e.g., happy \& surprised, sad \& angry) expressions based on face appearance for both in-the-lab \cite{lucey2010extended,pantic2005web,lyons1998coding} and in-the-wild environments \cite{Li_2017_CVPR,Dhall2011Static}. Recently, most fruitful algorithms resort to deep neural networks \cite{simonyan2014very,he2016deep} to learning powerful feature representation to promote the FER performance. Despite achieving impressive progress for the lab-controlled environments, it is still challenging and unsolved due to the complex variations in pose, illumination, and age, especially in the uncontrolled environments during the natural human robot interaction process.

\begin{figure}[!t]
   \centering
   \includegraphics[width=0.95\linewidth]{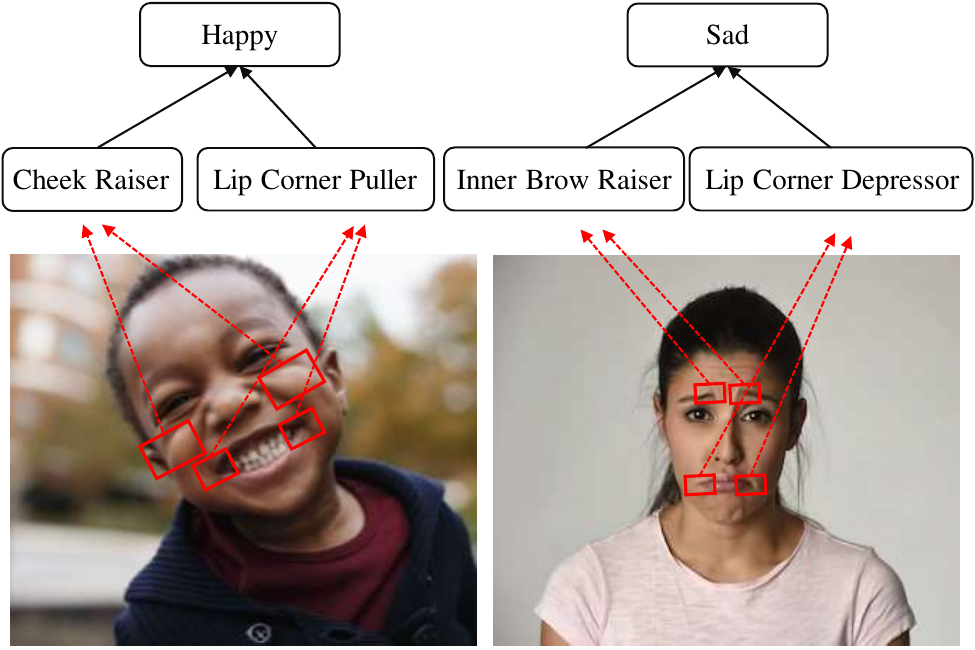}
   \caption{Two examples of the correlations between facial expression and action units.}
   \label{fig:motivation}
\end{figure}

According to the facial action coding system (FACS) \cite{friesen1978facial,tian2001recognizing}, facial action units (AUs) encode subtle facial appearances and changes, which have strong correlations with human expressions. For example as shown in Figure \ref{fig:motivation}, if a face image is detected to have the AUs of ``cheek raiser'' and ``lip corner puller'', it tends to be a ``happy'' face. In contrast, if the detected AUs are ``inner brow raiser'' and ``lip corner depressor'', it is more likely to be a ``sad'' face. Thus, automatically detecting the AUs and modeling their relationships with the expressions is essential to promote FER performance, especially to help distinguish uncertain or ambiguous expressions.

In this work, we aim to mine the AU features to enhance face image representation for more robust and accurate expression recognition. To achieve this end, two crucial challenges arise. First, most current FER datasets (e.g., RAF-DB \cite{Li_2017_CVPR}, SFEW2.0 \cite{Dhall2011Static}) do not have AU annotations, and it is very expensive and labor-consuming to annotate the AUs for these datasets. Thus, how to learn AU features without AU annotations is a key challenge. Second, there exist tens of AUs, and not all AUs are equally important for different expressions. How to adaptively select AU features to enhance image representation for each expression is another vital problem. 

To address these challenges, we explore exploiting the correlations among expressions and AUs to learn AU features in an unsupervised manner and adaptively select these features for feature enhancement by developing a novel AU-Expression Knowledge Constrained Representation Learning (AUE-CRL) framework. Specifically, there exist strong correlations among expressions and AUs. We first design a knowledge-guided AU representation learning module that leverages these correlations to covert expression labels to pseudo AU labels and utilizes pseudo labels to train the AU classifiers to obtain the AU features. As suggested in previous work, different AUs also have obvious co-occurrence dependencies, and these dependencies are important for selecting useful AUs. With the AU features, we further introduce a knowledge-guided attention module that learns to adaptively mine the most relevant AU features under the constraint of AU dependencies. In this way, the framework can automatically discover useful local facial behaviors to facilitate FER.

In summary, the contributions of this work are three-fold. First, we design a novel AU-Expression Knowledge Constrained Representation Learning (AUE-CRL) framework that exploits prior knowledge of AU-expression correlations to automatically discover useful AU features for feature enhancement to facilitate recognizing facial expression. Second, we propose to leverage the AU-expression correlations to guide learning AU features without AU annotations. Thus, the framework can be easily generalized to all of current FER datasets. Finally, we conduct extensive experiments on several large-scale in-the-wild datasets to demonstrate the effectiveness of the proposed framework, and carry out ablative studies to analyze the actual contribution of each key component. 

\begin{figure*}[!t]
   \centering
   \includegraphics[width=0.7\linewidth]{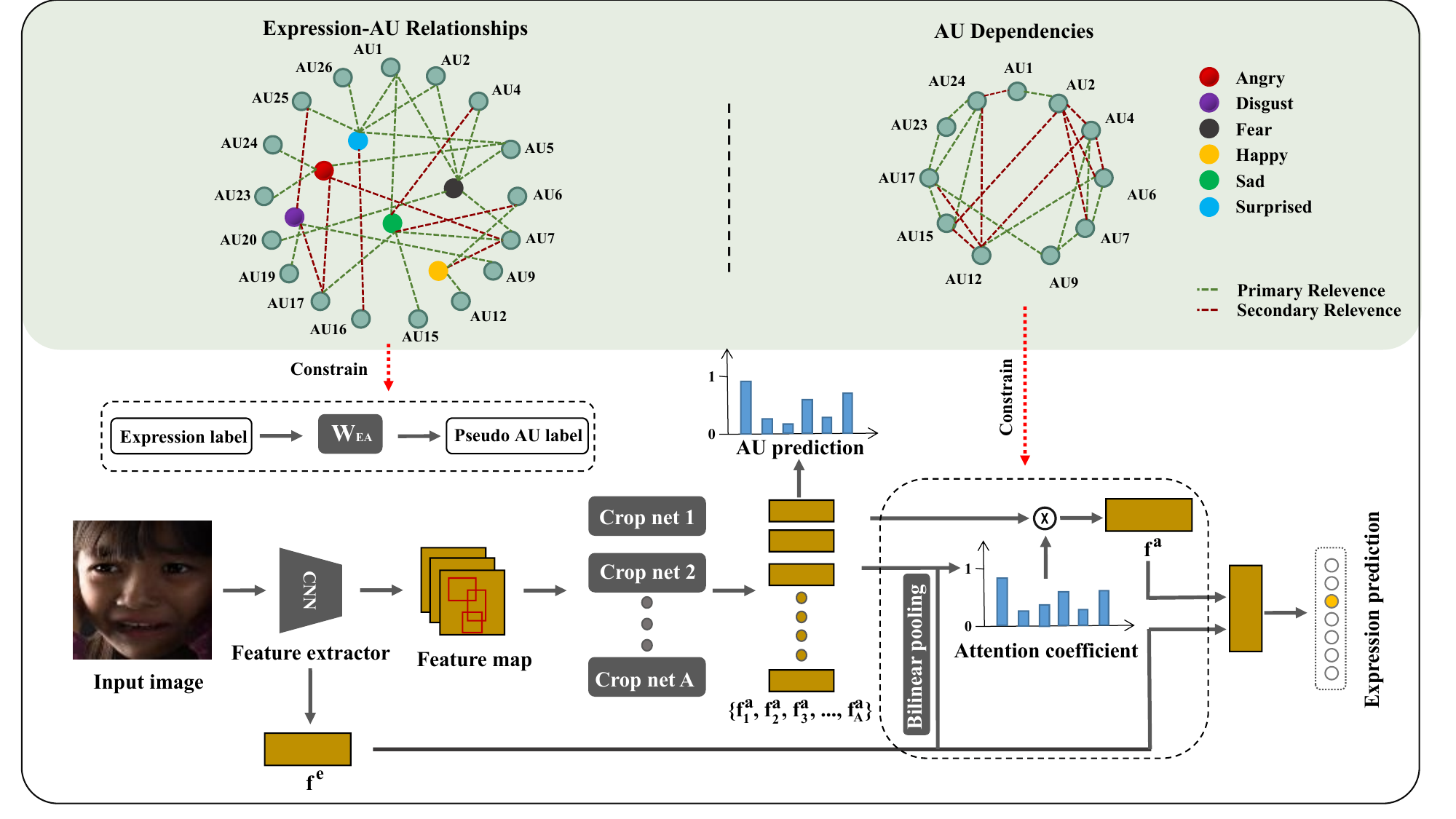}
   \caption{An illustration of the proposed framework. We exploit the relationships between AUs and expressions to guide learning AU representation in an unsupervised manner and the dependencies among AU to constrain the attention mechanism to better select useful AU representation to facilitate facial expression recognition. To keep this illustration concise, we only show part of relationships and dependencies.}
   \label{fig:framework}
\end{figure*}

\section{Related work}

In this section, we review the most related works about facial expression recognition and facial action unit detection.

\subsection{Facial Expression Recognition}

Previous works on facial expression recognition mainly focused on the basic categories (e.g., happy, angry, etc.) in which the data were collected by asking volunteers to make specific expressions in the constrained lab environments \cite{6222016,5771368}. Traditional methods primarily designed hand-crafted features (e.g., Local Binary Pattern (LBP) \cite{6222016}, Bag of Word (BoW) \cite{4813445}, and Histogram of Oriented Gradient (HoG) \cite{5771368}). These methods can achieve satisfactory performance in such constrained environments. Recently, there emerged various large-scale datasets, in which the data were captured in real-world scenarios \cite{li2019semantic,mollahosseini2017affectnet,liu2014facial}. Compared with previous in-the-lab settings, these datasets were even more challenging due to more variance in pose, illumination, etc. Previous hand-crafted features could hardly capture such variance and thus they worked quite poor on these datasets. To address this issue, recent works resorted to deep convolutional networks \cite{simonyan2014very,he2016deep} to learn more powerful feature representation for expression recognition. For example, Liu et al. \cite{liu2014facial} proposed a Boosted Deep Belief Network to learn feature that could characterize expression-related facial appearance/shape changes. Mollahosseini et al. \cite{mollahosseini2016going} designed deeper convolutional networks that built on inception module \cite{Szegedy_2015_CVPR} to extract more discriminative feature for recognition. Liu et al. \cite{10.1007/978-3-319-16817-3_10} exploited 3D Convolutional Neural Networks that was trained with deformable action parts constraints to learn discriminative part-based representation. Yu et al. \cite{Yu:2015:IBS:2818346.2830595} ensembled multiple deep Convolutional Neural Networks to further improve recognition accuracy. Different from these works, our method introduces the relationships between AU-expression and relationship among AUs to mine information of AUs to help capture local facial behaviors to facilitate facial expression recognition. 

The proposed method is also related to some recent works that also exploit prior knowledge to facilitate visual reasoning \cite{xie2020adversarial,chen2020cross,chen2020knowledge1,chen2019learning,chen2019knowledge,wang2018deep}. For example, Chen et al. introduce the co-occurrence correlations among different categories to help better recognize multiple semantic objects \cite{chen2019learning,chen2020knowledge1}. Generally, these methods represent prior knowledge in the form of graphs and introduce the graph neural networks \cite{li2016gated,kipf2017semi} for message propagation to learn contextualized feature representations. Different from these works, we introduce the relationships between AU-expression and among AUs to guide generating pseudos AU labels for AU representation learning and selecting useful AUs for facial expression recognition. 

\subsection{Action Unit Recognition}

Facial action units are defined to describe facial muscle movements according to \cite{PE1978FACS}, and detecting action unit is helpful for expression and emotion understanding. Previous works \cite{savran2012regression,mahoor2009framework} leveraged traditional shadow models (e.g., SVM and SVR) to solve this task. For example, Mahoor et al. \cite{mahoor2009framework} projected high-dimensional facial images into a low-dimensional space via the spectral regression technique and adopted SVM classifier to predict the AU intensity. Inspired by recent advance of deep neural network on vision tasks, current works \cite{batista2017aumpnet,zhou2017pose,walecki2017deep,gudi2015deep} also designed deep model for action unit recognition. Gudi et al. \cite{gudi2015deep} designed a simple seven-layer network to estimate both occurrences and intensities of the AUs. Furthermore, Walecki et al. \cite{walecki2017deep} adopted the conditional random field (CRF) to encode AU dependencies and combined it with deep neural networks to improve recognition.

\section{AUE-CRL Framework}

\subsection{Overview}

This proposed AUE-CRL framework explores mining useful local AU information to enhance image representation learning under the guidance of the correlations among AUs and expressions. It mainly consists of modules, i.e., feature extractors, knowledge-guided AU representation learning (KGAURL) module, and knowledge-constrained AU selection (KCAUS) module. Taking an input image $I$, the feature extractor generates multi-layer feature maps, and then it fuses these feature maps and performs global average pooling to obtain global expression feature vector $\textbf{f}^e \in \mathcal{R}^{d_e}$. The feature extractor also fuses the feature maps from multiple layers inversely to obtain a set of feature maps with relative large size and leverage a crop network to extract feature for each AU, i.e., $\{\textbf{f}^a_1, \textbf{f}^a_2, \dots, \textbf{f}^a_A\}$. Here, $\textbf{f}^a_i \in \mathcal{R}^{d_a}$ is the feature vector of the $i$-th AU and $A$ is the AU number. Then, the KGAURL module converts the expression labels to pseudo AU labels based on the AU-expression correlations. The pseudo labels are then used to supervised AU classifier training and thus it can learn AU features without any additional AU annotations. Finally, the KCAUS exploits an attention mechanism to automatically discover useful AU features under the constraint of AU dependencies, and aggregate these features with the global expression feature vector for expression prediction. An overall framework is illustrated in Figure \ref{fig:framework}. 

\subsection{Knowledge-Guided AU Representation Learning}

Current algorithms \cite{li2019semantic,gudi2015deep} mainly resort to deep neural networks \cite{he2016deep,simonyan2014very,chen2018learning} to learn AU representations, but they requires a large amount of ground truth annotations to ensure the discriminative and generalization abilities. However, current datasets with AU annotations are mainly captured in the constrained in-the-lab environment and cover very few subjects, e.g., 27 and 41 subjects on the DISFA \cite{mavadati2013disfa} and BP4D \cite{zhang2014bp4d} datasets. Features trained using these datasets can hardly generalize to other environments and subjects, especially to the in-the-wild settings. On the other hand, current FER datasets lack the AU annotations, and it is expensive and unpractical to add the AU annotations for these datasets. In this work, we design a knowledge-guided module that exploits the correlations between AU and expression to generate pseudo AU labels, and thus the proposed framework can learn AU representation without incurring additional annotations. 

As suggested in previous literatures \cite{PE1978FACS}, each expression is relevant with several AUs, and the relevance can be further divided into primary and secondary ones. More concretely, if a face image is annotated with a specific expression, it tends to have the corresponding primary AUs with high probabilities, secondary AUs with middle probabilities, and other AUs with low probabilities. According to these relationships, we can build a correlation matrix $\widehat{\mathbf{W}}_{EA}$, where $E$ and $A$ denote the expression and AU number, respectively. The value $\hat{w}_{ea}$ denote the prior relevant probability between expression $e$ and AU $a$. It is assigned with a large value if they are primarily relevant, a middle value if they are secondarily relevant, and a small value otherwise. Given a sample with expression annotations of $\hat{\mathbf{p}}_e=\{\hat{p}_{e0}, \hat{p}_{e1}, \dots, \hat{p}_{e{E-1}}\}$, it is intuitive to produce the pseudo AU labels by $\hat{\mathbf{p}}_e\widehat{\mathbf{W}}_{EA}$. However, the matrix merely defines three level correlations, which is very cursory. It is desirable to exploit finer-grained correlations so as to obtain more precise pseudo AU labels. In this work, we use a learnable correlation matrix $\mathbf{W}_{EA}$ to generate the pseudo AU labels by 
\begin{equation}
  \hat{\mathbf{p}}_a=\hat{\mathbf{p}}_e\mathbf{W}_{EA},
\end{equation}
We apply the simple linear function to map the AU features $\{\textbf{f}^a_1, \textbf{f}^a_2, \dots, \textbf{f}^a_A\}$ to the predicted AU labels
\begin{equation}
  \mathbf{p}_a=\{\mathbf{w}_1\textbf{f}^a_1, \mathbf{w}_2\textbf{f}^a_2, \dots, \mathbf{w}_A\textbf{f}^a_A\},
\end{equation}
where $\mathbf{w}_a$ is a learnable weight vector. During training, we define a mean square error loss between the pseudo and the predicted labels, and a regularization loss between the learned and prior correlation matrices, formulated as
\begin{equation}
  \mathcal{L}_{au}=||\mathbf{p}_a-\hat{\mathbf{p}}_a||_2^2+\lambda||\mathbf{W}_{EA}-\widehat{\mathbf{W}}_{EA}||_2^2.
\end{equation}

In this way, we can learn finer-grained correlation and simultaneously exploit prior correlations to promote generating more precise AU labels. $\lambda$ is a balance parameter and it is set to 1.0 in the experiments.

\subsection{Knowledge-Constrained AU Selection}
In this subsection, we introduce the knowledge-constrained attention mechanism that learns to adaptively select useful AU features and fuses these features to enhance image representation. It computes a correlation coefficient for each AU, performs weighted average to obtain a merged AU feature, and concatenates it with expression feature to obtain the final image representation.

Specifically, we first fuse the expression features with each AU feature using the low-rank bilinear pooling \cite{Kim2016Hadamard} to compute a coefficient
\begin{equation}
  \hat{w}_i =\mathbf{P}^{T}(\tanh(\mathbf{U}^{T}\textbf{f}^e)\odot\tanh(\mathbf{V}^{T}\textbf{f}^a_{i})+\mathbf{b}))
\end{equation}
that denotes the importance of AU $i$ for expression recognition. Here, we use low-rank bilinear pooling \cite{Kim2016Hadamard} as it is effective for feature fusion. In the equation, $\tanh(\cdot)$ is the hyperbolic tangent function and $\odot$ is the element-wise product operation. $U \in R^{d_e \times d}$, $V \in R^{d_a \times d}$, $P \in R^{d \times 1}$ are the learnable parameter matrixes. To make the coefficients easily comparable across different samples, we normalize the coefficients over all AUs using a softmax function
\begin{equation}
  w_i =\frac{\hat{w}_i}{\sum_{j}\hat{w}_j}.
\end{equation}
Then, we perform weighted average over all AUs to obtain the AU features
\begin{equation}
  \mathbf{f}^a=\sum_{i}w_i\textbf{f}^a_i.
\end{equation}
Finally, we concatenate $\mathbf{f}^a$ with $\mathbf{f}^e$ for expression prediction
\begin{equation}
  \mathbf{p}^e=f([\mathbf{f}^a, \mathbf{f}^e]).
\end{equation}

\subsection{Knowledge-Regularized Training Loss}

As suggested in previous literatures \cite{friesen1978facial}, there exists strong co-occurrence dependencies among different AUs. In other words, some AUs co-occur frequently while some AUs are mutually exclusive. For example, the AU \emph{inner brow raiser} always co-occurs with \emph{outer brow raiser} as they are both controlled by the muscle group of \emph{Occipito franontanlis}. In contrast, the AU \emph{lip corner puller} hardly co-exists with \emph{lip corner depressor} as the corresponding controlled muscle group can not co-activate. It is expected and natural that the learned attentional coefficients should match such dependencies, and thus we introduce these prior dependencies as a regularization term during training. 

Inspired by previous work \cite{zhang2018classifier}, we consider the pair-wise dependencies that include positive and negative correlations to define the regularization term. Here, we consider the AU $i$ exists if its attention coefficient is higher than 0.5. We denote it as  $i_1$ if it exists and denote as $i_0$ otherwise. For the AU $i$ and $j$ with positive correlation, it is expected
\begin{equation}
   \begin{split}
   p(i_1|j_1)&>p(i_0|j_1) \\
   p(i_1|j_1)&>p(i_1|j_0)
   \end{split}
\end{equation}
which can be transformed to the equivalent formulation
\begin{equation}
   \begin{split}
   p(i_1,j_1)&>p(i_0,j_1) \\
   p(i_1,j_1)&>p(i_1)p(j_1) \\
   \end{split}
\end{equation}
Accordingly, we can define the regularization term that constrains the positive correlation as 
\begin{equation}
  \begin{split}
  \ell_p=&\sum_{i,j \in S_p}\max(p(i_1)p(j_1)-p(i_1,j_1), 0)\\ 
  & + \sum_{i,j \in S_p}\max(p(i_1,j_0)-p(i_1,j_1), 0)\\
  & + \sum_{i,j \in S_p}\max(p(i_0,j_1)-p(i_1,j_1), 0).
  \end{split}
\end{equation} 
where $S_p$ is the set of positive AU pairs. Similarly, the regularization term for negative correlation constraint can be defined as
\begin{equation}
  \begin{split}
  \ell_n=&\sum_{i,j \in S_n}\max(p(i_1,j_1)-p(i_1)p(j_1), 0)\\
  & + \sum_{i,j \in S_n}\max(p(i_1,j_1)-p(i_1,j_0), 0)\\
  & + \sum_{i,j \in S_n}\max(p(i_1,j_1)-p(i_0,j_1), 0).
  \end{split}
\end{equation} 
where $S_n$ is the set of negative AU pairs.

During training, we use the cross entropy loss, which is denoted by $\ell_c$, to train the expression classifier, and thus the loss can be defined as 
\begin{equation}
  \mathcal{L}_e=\ell_c+\alpha(\ell_p+\ell_n),
\end{equation}
where $\alpha$ is the balance parameter and it is set to 0.5 in the experiments.
 
\subsection{Implementation Details}

\subsubsection{Network architecture}

We select ResNet-101 \cite{he2016resnet} as the backbone network for feature extraction, which consists of four block layers. Given an input image of size 224 $\times$ 224, we can obtain feature maps of size 56 $\times$ 56 $\times$ 256 from first layer, feature maps of size 28 $\times$ 28 $\times$ 512 from second layer, feature maps of size 14 $\times$ 14 $\times$ 1024 from third layer and feature maps of size 7 $\times$ 7 $\times$ 2048 from last layer. For holistic expression feature, we downsample the feature maps from the first, second, third layer of backbone network to the size of feature maps from last layer by max pooling, then concatenate these four feature maps, and perform global average pooling to obtain a 3840-dimensional vector. For AU feature, we inversely upsample the feature maps to the size of feature maps from previous layer by deconvolution, starting from feature maps from last layer to the end of second layer, and concatenate each upsampled feature maps with original feature maps, and upsample it again by deconvolution. At end, we obtain a feature map of size 56 $\times$ 56 $\times$ 256, which is used for cropping feature from corresponding location to obtain the corresponding AU feature by using crop net.

In crop net, we first use MTCNN \cite{Zhang2016MTCNN} to get facial landmarks of input image and crop the corresponding region for each AU on feature maps by using code of \cite{li2019semantic}, and pass it through a convolutional layer and a fully connected layer, whose parameters are not shared for each AU, to obtain a 512-dimensional vector.

\subsubsection{Training details}

To obtain more stable experiment results, we adopt three-stages training process. In the first stage, we train the backbone and expression classifier with the cross-entropy loss using stochastic gradient descent(SGD) with an initial learning rate of 0.01, a momentum of 0.9, and a weight decay of 0. In the second stage, we fix the parameters of backbone, and train crop net and AU classifier with mean-square error loss and loss $\mathcal{L}_{au}$ defined by the formula (3) using stochastic gradient descent(SGD) with an initial learning rate of 0.001, a momentum of 0.9, and a weight decay of 0. And in third stage, we fix the parameters of backbone, crop net and AU classifier, and train expression classifier with loss $\mathcal{L}_{e}$ defined by the formula (12) using stochastic gradient descent(SGD) with an initial learning rate of 0.0001, a momentum of 0.9, and a weight decay of 0.

\begin{table*}
  \centering
  \begin{tabular}{c|ccccccc|c}
  \toprule
  \centering  Methods & Angry & Disgust & Fear & Happy & Neutral & Sad & Surprised & Ave. acc \\
  \hline
  \centering  DCNN-DA \cite{kollias2020generating} & 78.4 & 64.4 & 62.2 & 91.1 & 80.6 & 81.2 & 84.5 & 77.5 \\
  \centering  WSLGRN \cite{Zhang_2020_IJCAI} & 75.3 & 56.9 & 63.5 & 93.8 & 85.4 & 83.5 & 85.4 & 77.7 \\
  \centering  CP \cite{Acharya_2018_CVPR_Workshops} & 80.0 & 61.0 & 61.0 & 93.0 & \textbf{89.0} & \textbf{86.0} & 86.0 & 79.4 \\
  \centering  CompactDLM \cite{Kuo_2018_CVPR_Workshops} & 74.5 & 67.6 & 46.9 & 82.3 & 59.1 & 58.0 & 84.6 & 67.6 \\
  \centering  FSN \cite{Zhao_2018_BMVC} & 72.8 & 46.9 & 56.8 & 90.5 & 76.9 & 81.6 & 81.8 & 72.5 \\
  \centering  DLP-CNN \cite{Li_2017_CVPR} & 71.6 & 52.2 & 62.2 & 92.8 & 80.3 & 80.1 & 81.2 & 74.2 \\
  \centering  MRE-CNN \cite{fan2018multi}  & \textbf{84.0} & 57.5 & 60.8 & 88.8 & 80.2 & 79.9 & 86.0 & 76.7 \\
  \hline
  \centering  Ours & 80.5 & \textbf{67.6} & \textbf{68.9} & \textbf{94.1} & 85.8 & 83.6 & \textbf{86.4} & \textbf{81.0} \\
  \bottomrule
  \end{tabular}
  \vspace{2pt}
  \caption{Performance of our proposed method and current existing state-of-the-art competitors for recognizing the basic expressions on the RAF-DB dataset.}
  \label{tab:RAF-basic-acc}
\end{table*}

\begin{table}
  \centering
  \begin{tabular}{ccccc}
  \toprule
  \centering  Methods & BaseDCNN \cite{Li_2017_CVPR} & Center Loss \cite{Li_2017_CVPR} & DLP-CNN \cite{Li_2017_CVPR} & Ours \\
  \hline
  \centering  Ave. acc & 40.2 & 40.0 & 44.6 & \textbf{51.1}\\
  \bottomrule
  \end{tabular}
  \vspace{2pt}
  \caption{Performance of our proposed method and current existing state-of-the-art competitors for recognizing the compound expressions on the RAF-DB dataset.}
  \label{tab:RAF-compound-acc}
\end{table}

\begin{table*}
  \centering
  \begin{tabular}{c|ccccccc|c}
  \toprule
  \centering  Methods & Angry & Disgust & Fear & Happy & Neutral & Sad & Surprised & Ave. acc \\
  \hline
  \centering  CP \cite{Acharya_2018_CVPR_Workshops} & 66.0 & 0.0 & 14.0 & \textbf{90.0} & 86.0 & \textbf{66.0} & 29.0 & 50.1 \\
  \centering  DLP-CNN \cite{Li_2017_CVPR} & - & - & - & - & - & - & - & 51.1\\
  \centering IA-CNN \cite{meng2017identity} & 70.7 & 0.0 & 8.9 & 70.4 & 60.3 & 58.8 & 28.9 & 42.6 \\
  \centering  IL \cite{cai2018island} & 61.0 & 0.0 & 6.4 & 89.0 & 66.2 & 48.0 & 33.3 & 43.4\\
  \hline
  \centering Ours & \textbf{75.3} & \textbf{17.4} & \textbf{25.5} & 86.3 & \textbf{72.1} & 50.7 & \textbf{42.1} & \textbf{52.8} \\
  \bottomrule
  \end{tabular}
  \vspace{2pt}
  \caption{Performance of our proposed method and current existing state-of-the-art competitors recognizing the basic expressions on the SFEW2.0 dataset. - denotes corresponding result is not provided.}
  \label{tab:SFEW-basic-acc}
\end{table*} 

\begin{table*}
  \centering
  \begin{tabular}{c|ccccccc|c}
  \toprule
  \centering  Methods & Surprised & Fear & Disgust & Happy & Sad & Angry & Neutral & Ave. acc \\
  \hline
  \centering Baseline & 85.5 & 68.9 & 60.1 & 94.1 & 85.3 & 81.8 & 83.3 & 79.9 \\
  \centering Ours w/o KGAURL & 88.3 & 63.5 & 64.2 & 92.7 & 85.7 & 79.9 & 87.3 & 80.2 \\
  \centering Ours w/o Attention & 88.0 & 67.6 & 60.1 & 94.1 & 84.0 & 82.5 & 85.0 & 80.2 \\
  \centering Ours w/o AU-Independent & 86.1 & 67.6 & 65.5 & 94.2 & 84.6 & 79.9 & 85.9 & 80.6 \\
  \centering  Ours & 86.4 & 68.9 & 67.6 & 94.1 & 83.6 & 80.5 & 85.8 & 81.0 \\
  \bottomrule
  \end{tabular}
  \vspace{2pt}
  \caption{Performance of our method (Ours), our method without knowledge-guided AU representation Learning (Ours w/o KGAURL), Our method without attention mechanism for AU representaion selection (Ours w/o Attention), Our method without AU-indenpent constrain (Ours w/o AU-Indenpendent), and the baseline resnet 101 (Baseline) on the RAF-DB dataset.}
  \label{tab:ablative}
\end{table*}

\section{Experiments}

\subsection{Datasets}

The existing datasets of facial expression can be divided into two main categories according to its collecting environment. Over quite a long period there only exist datasets collected in the lab-controlled conditions with limited size. Recently, some comparatively large-scale datasets that reflect real-world scenarios are released to promote the research. Meanwhile, the types of expressions are expanded with compound expressions that can be constructed by the combination of basic expression.

We chose two challenging in-the-wild datasets to evaluate the performance of our method. The Real-world Affective Face Database (RAF-DB) \cite{Li_2017_CVPR} and the Static Facial Expressions in the Wild (SFEW2.0) \cite{Dhall2011Static} datasets.

RAF-DB \cite{Li_2017_CVPR} contains 29,672 highly diverse facial images from thousands of individuals that were also collected from the Internet. With manually crowd-sourced annotation and reliable estimation, seven basic and eleven compound emotion labels are provided for the samples. Specifically, 15,339 images from the basic emotion set are divided into two groups (12,271 training samples and 3,068 testing samples) for evaluation.

SFEW2.0 \cite{Dhall2011Static} is an in-the-wild dataset collected from different films with spontaneous expressions, various head poses, age ranges, occlusions and illuminations. This dataset is divided into training, validation, and test sets, with 958, 436, and 372 samples, respectively.

\subsection{Comparison with State-of-the-Arts}

In this subsection, we present the performance comparisons with current state-of-the-art methods to evaluate the superiority of our proposed method.

\subsubsection{Performance on RAF-DB}

RAF-DB is a challenging in-the-wild dataset that is widely used for evaluating facial expression recognition. In this part, we compare our method with current state-of-the-art competitors, including Deep Neural Network Augmentation(DCNN-DA) ~\cite{kollias2020generating}, Weakly Supervised Local-Global Relation Network (WSLGRN) ~\cite{Zhang_2020_IJCAI}, Covariance Pooling (CP) \cite{Acharya_2018_CVPR_Workshops}, Compact Deep Learning Model(CompactDLM) ~\cite{Kuo_2018_CVPR_Workshops}, Feature Selection Network(FSN) ~\cite{Zhao_2018_BMVC}, Deep Locality-Preserving Learning (DLP-CNN) ~\cite{Li_2017_CVPR}, and Multi-Region Ensemble CNN (MRE-CNN) ~\cite{fan2018multi}. 

We first present the accuracy of each basic expression and the average accuracy over all expressions in Table \ref{tab:RAF-basic-acc}. As shown, our method achieves the best performance compared with existing competitors, i.e., improving the average accuracy from 79.4\% to 81.0\%. In addition, our method obtains better accuracy for most basic expression, especially for those that are difficult to recognize. For example, current best accuracy for the expression ``Fear'' is 63.5\%. Our method improves the accuracy to 68.9\%, with a relative improvement of 8.50\%. One possible reason is integrating information of facial AU can well capture local discriminative feature and help to distinguish uncertain and ambiguous expression.

Except for the basic expression, RAF-DB contains another sub-set in which each face image is annotated with a compound expression. A compound expression usually contains two or more basic expressions. For example, a person may be happy and simultaneously surprised. Obvious, this is an even more difficult task as it needs to recognize multiple expression patterns, which depends more on local discriminative feature mining. Here, we compare our method with BaseDCNN \cite{Li_2017_CVPR}, Center Loss ~\cite{Li_2017_CVPR}, and Deep Locality-Preserving Learning (DLP-CNN) ~\cite{Li_2017_CVPR}. As shown in Table \ref{tab:RAF-compound-acc}, our method outperforms current state-of-the-art competitors by a sizable margin, i.e., an improvement of 6.50\% in average accuracy.

\subsubsection{Performance on SFEW2.0}

SFEW2.0 is an even more challenging dataset, and there are also some works that conduct experiments on this dataset. Here, we compare with our proposed method with the following works: Covariance Pooling (CP) \cite{Acharya_2018_CVPR_Workshops}, Deep Locality-Preserving Learning (DLP-CNN) \cite{Li_2017_CVPR}, Identity-Aware Convolutional Neural Network (IA-CNN) \cite{meng2017identity} , and Island Loss (IL) \cite{cai2018island}. 

The accuracy of each basic expression and average accuracy overall all expressions are presented in Table \ref{tab:SFEW-basic-acc}. Our method obtains an average accuracy of 52.8\%, improving that of the previous best method by 1.7\%. Similar to results on RAF, existing methods perform extremely poor for the expressions ``Disgust'' and ``Fear'', e.g., accuracies of 0.0\% and 14.0\% for these two expressions. Our methods improve the accuracies to 17.4\% and 25.5\%. These comparisons again demonstrate the superiority of our proposed method in ambiguous expression recognition.

\subsection{Ablation Study}

In this subsection, we conduct comprehensive ablation studies to discuss and analyze the contribution of each component and obtain a more thorough understanding of the framework.

\subsubsection{Analysis of AUE-CRL framework}

The core contribution of the proposed framework is mining useful local AU information to enhance image representation. To analyze the contribution of this framework, we compare AUE-CRL framework with the ResNet-101 baseline. The experiment is conducted on the RAF-DB dataset and the results are presented in Table \ref{tab:ablative}. As shown, the average accuracy drops from 81.0\% to 79.9\%. It is worth noting that, compared to AUE-CRL framework, the accuracy of baseline drops significantly on expression "Disgust", which proves AUE-CRL framework is effective to distinguish uncertain and ambiguous expression.

\subsubsection{Analysis of knowledge-guided AU representation learning}

As suggested above, we leverage the relationship of AU and expression to guide learning AU representation, which can help get rid of heavy AU annotations and guide learning domain-adaptive representation. To  analyze the effect of the relationship of AU and expression, we remove the AU-expression regularization loss and use AU annotation of the BP4D dataset \cite{zhang2014bp4d} to train AU classifiers. We conduct the comparison on the RAF-DB dataset and present the results on Table \ref{tab:ablative}. Although this method adopts ground truth AU annotations, it performs inferior compared with our knowledge-guided AU representation learning, i.e., decreasing the accuracy from 81.0\% to 80.2\%. 

Indeed, the images of BP4D cover merely 41 subjects and they are captured in the constrained lab environment. Thus, representation learned on such a dataset can hardly generalize to other environments. In contrast, our proposed knowledge-guided AU representation learning enables training on the target dataset and tends to learn domain-adaptive AU representation, leading to better performance.

\subsubsection{Analysis of knowledge-constrained AU selection}

In this work, we introduce a knowledge-constrained attention mechanism to adaptively select useful AU for expression representation enhancement. To analyze its contribution, we remove this component, simply perform average pooling over all action unit to obtain AU representation and concatenate it with expression feature for expression recognition. We find the average accuracy drops to 80.2\% and accuracy drops significantly on expression "Disgust", which suggests that the attention mechanism can help mine useful AU information to facilitate expression recognition and play a great role in distinguish uncertain and ambiguous expression. 

To better select useful AUs, we introduce prior knowledge of the relationship among AUs as a constraint term during training. Here, we remove this constraint to analyze its contribution. As shown in Table \ref{tab:ablative}, we find the average accuracy is 80.6\% on the RAF-DB dataset, which is better than that without the attention mechanism but still worse than our method.

\section{Conclusion}

In this paper, we propose an AU-Expression Knowledge Constrained Representation Learning framework that exploits prior knowledge to help mining AU information to promote facial expression recognition. It first leverages relationships between AUs and expressions to guide learning domain-adaptive AU representation without any additional AU annotations. Then, it introduces an attentional mechanism to adaptively select useful AU representation under the constraint of the dependencies among AUs. We conduct an experiment on two in-the-wild datasets and show that our method outperforms current state-of-the-art competitors by a sizable margin. 

\bibliographystyle{./IEEEtran}  
\bibliography{./reference}

\addtolength{\textheight}{-12cm}  

\end{document}